\documentclass[journal]{IEEEtran}

\usepackage{mathtools}
\usepackage{stfloats}

\usepackage{multirow}
\usepackage{color, soul}

\usepackage{tikz}
\usepackage{textcomp}
\usepackage{hyperref}
\usepackage{lipsum}

\begin{document}

\title{Curse of Small Sample Size in Forecasting of the Active Cases in COVID-19 Outbreak}

\author{\IEEEauthorblockN{\IEEEauthorrefmark{1}Mert~Nakıp,~\IEEEmembership{Student Member,~IEEE,}
		\IEEEauthorrefmark{2}Onur~Çopur, \IEEEauthorrefmark{3}Cüneyt~Güzeliş 
		\thanks{Mert~Nakıp (Corresponding Author) is with Institute of Theoretical and Applied Informatics, Polish Academy of Sciences (PAN), ul. Baltycka 5, 44100 Gliwice, Poland. Email: mnakip@iitis.pl}
		\thanks{Cüneyt~Güzeliş is with the Department of Electrical-Electronics Engineering at Yaşar University, Izmir, Turkey. Email: cuneyt.guzelis@yasar.edu.tr.}
		\thanks{Onur~Çopur is a student at the Department of Statistics at Sapienza Universiy of Rome, Rome, Italy. Email: copur.1891194@studenti.uniroma1.it}}
}

\maketitle

%\copyrightnotice

\begin{abstract}

The COVID-19 pandemic has affected almost all countries in the world in the first half of 2020. During this time, a massive number of attempts on the predictions of the number of cases and the other future trends of this pandemic have been made. However, they fail to predict, in a reliable way, the medium and long term evolution of fundamental features of COVID-19 outbreak within acceptable accuracy. This paper gives an explanation for the failure of machine learning models in this particular forecasting problem. The paper shows that simple linear regression models provide high prediction accuracy values reliably but only for a 2-weeks period and that relatively complex machine learning models, which have the potential of learning long term predictions with low errors, cannot achieve to obtain good predictions with possessing a high generalization ability. It is suggested in the paper that the lack of a sufficient number of samples is the source of low prediction performance of the forecasting models. The reliability of the forecasting results about the active cases is measured in terms of the cross-validation prediction errors, which are used as expectations for the generalization errors of the forecasters. To exploit the information, which is of most relevant with the active cases, we perform feature selection over a variety of variables such as the numbers of active cases, deaths, recoveries, and people per kilometer square. We apply different feature selection methods, namely the Pairwise Correlation, Recursive Feature Selection, and feature selection by using the Lasso regression and compare them to each other and also with the models not employing any feature selection. Furthermore, we compare Linear Regression, Multi-Layer Perceptron, and Long-Short Term Memory models each of which is used for prediction active cases together with the mentioned feature selection methods. Our results show that the accurate forecasting of the active cases with high generalization ability is possible up to 3 days only because of the small sample size of COVID-19 data. We observe that the linear regression model has much better prediction performance with high generalization ability as compared to the complex models but, as expected, its performance decays sharply for more than 14-days prediction horizons. 

\end{abstract}


\IEEEpeerreviewmaketitle

\section{Introduction}
\label{sec:Introduction}

Since the first COVID-19 case confirmed on December 2019 in Wuhan, the COVID-19 outbreak has been spreading with acceleration all around the world. According to rapid spread of this pandemic, in the most of the countries, the first concern was that the medical facilities may not be sufficient to handle with the massive number of patients. To plan necessary actions such as increasing the facilities or taking preventive decisions to flatten the curve of daily cases, the determination of the future pattern of the active cases has become one of the most important issues. So, many studies on the forecasting of the number of active cases have been published in the literature \cite{benvenuto2020application, anastassopoulou2020data, petropoulos2020forecasting}. Although there are many valuable results in the published works, some of the publications optimistically make long term predictions for the pandemic \cite{anastassopoulou2020data, yang2020modified}. Furthermore, in most of the works, the test performance of the forecasting results has not been demonstrated well due to the restricted size of the available time series data covering several months only \cite{benvenuto2020application, anastassopoulou2020data, yang2020modified}. 

In this study, we perform an analysis on the generalization ability of the data-dependent forecasters to explain why forecasting models used for determining the trend of COVID-19 cases possess poor medium and long term prediction performances in the special case of machine learning models. For this purpose, a forecasting system that consists of a feature selection module and a machine learning based forecasting module is designed and implemented. In the early phase of our studies, we observed that such an architecture provides the best forecasting performance among the considered models including the standard architectures of Linear Regression (LR), Multi-Layer Perceptron (MLP), and Long-Short Term Memory (LSTM) state-of-the-art models with or without feature selection. The rational behind the choice of these models relies on the following three facts: 1) LR is a linear static model which is the least complex architecture, so possessing the high generalization ability \cite{yan2009linear}. 2) MLP is a nonlinear static neural network model which has universal function approximation property and can be said to be the most widely used neural network model with producing successful results in many applications \cite{hastie2009elements}. 3) LSTM is a recurrent neural network model which is capable of approximating to the nonlinear dynamics and has proved itself as the best model in many challenging applications requiring to capture the temporal relations hidden in inherently nonlinear dynamics \cite{sak2014long}. In order to determine the best performance provided by this feature selection based forecasting model in terms of the cross-validation error, we trained and tested all of the feature selection and forecaster pair combinations: For the feature selection module, No Feature Selection (No FS), iterative feature selection based on the Pairwise Correlation (PCorr), Recursive Feature Selection (RFS), and feature selection by using the Lasso regression (Lasso, in short) \cite{muthukrishnan2016lasso} are used. For the machine learning module, LR, MLP, and LSTM are chosen as the forecasters to process the selected features. 

\subsection{Relationship to the State of the Art}
Now, we present the relationship between our study and the works that aim to forecast the active cases in COVID-19 outbreak. According to the best of the authors' knowledge, with respect to the method of forecasting, we classify the studies that forecast the active cases in COVID-19 outbreak into 3 categories as follows: (1) SIR (Susceptible, Infected, Recovered) family \cite{ndiaye2020analysis}; (2) statistical time series analysis methods (for example, Auto-Regressive Integrated Moving Average (ARIMA)) \cite{benvenuto2020application}; (3) machine learning methods \cite{pereira2020forecasting, rizk2020COVID, villalobos2020using}.

The SIR model is a dynamical compartmental model for describing the time evolution of a disease transmitted from human to human within a population by a set of nonlinear ordinary differential equations. In the SIR model, the total population is assumed to be constant and divided into the following classes: Susceptible (S), Infected (I), and Recovered (R) \cite{ndiaye2020analysis}. The works in \cite{ndiaye2020analysis, anastassopoulou2020data, roda2020difficult} use the SIR model as the estimator for the number of active cases. In \cite{ndiaye2020analysis}, the authors focus on forecasting the active cases for all countries, where forecasting horizon is 5 days. The study in \cite{anastassopoulou2020data}, SIRD (Susceptible, Infectious, Recovered, Dead) method is used to predict the active cases under 3 different scenarios for only the Hubei province of China. Furthermore, in \cite{roda2020difficult}, the authors compare SIR with SEIR (Susceptible, Exposed, Infectious, Recovered) considering only active cases in Wuhan. They show that the SIR model performs much better than the SEIR model in representing the information contained in the confirmed-case data. This indicates that predictions using more complex SIR-like dynamical models may not be reliable in comparison to the ones using simpler SIR-like models. On the other hand, although SIR-like models explains rise-and-fall nature of growth of the pandemic, they fail to capture the peak and the whole time-evolution of the disease within a reasonable accuracy due to the sensitive dependence of the time waveform of the solutions to the SIR differential equations on model parameters such as the average number of contacts per person per time. It should be noted that SIR-like models could be used for an accurate prediction of active cases only when highly accurate parameter estimations are achievable depending on the real data.   

Besides SIR, statistical time series analysis methods are also used to predict the active cases in COVID-19 outbreak. For the COVID-19 outbreak in Italy, the works forecast the active cases by using ARIMA which is a linear time-invariant dynamical model with stochastic input\cite{ceylan2020estimation, ribeiro2020short, kumar2020forecasting}, for 2 days \cite{benvenuto2020application} and Seasonal ARIMA (SARIMA) for 60 days \cite{chintalapudi2020COVID}. In \cite{petropoulos2020forecasting}, the exponential smoothing based models are used to predict future of the cumulative number of cases. The results of the works in this category show that these forecasters make the prediction similar to the trend of the past data. Thus, although those forecasters are able to capture the increasing trend of the active cases until the peak point, they are not capable of determining the whole time evolution of the disease. 

There are a few works that uses machine learning methods in order to forecast the active cases in the COVID-19 pandemic. In the works \cite{pereira2020forecasting, zandavi2020forecasting, pal2020neural, vadyala2020prediction}, the LSTM based models are used to forecast the future of the pandemic by training the model for the past COVID-19 data for each of the selected countries. Similar to this work, the MLP based models in \cite{rizk2020COVID, tamang2020forecasting}, support vector machine models in \cite{yadav2020analysis} and the logistic regression in \cite{villalobos2020using} were trained and then tested on the COVID-19 data of each country that was selected for test. In \cite{yang2020modified}, the authors took into account the problem of the small sample size for the COVID-19 pandemic and trained their model on the 2003 SARS corona virus outbreak data. The proposed results by the works in this category show that the forecasting models perform well with respect to the error metric measurements; however, the graphs show that the models are not able to capture either the peak days and the values or the non-increasing parts of the time series. 

The rest of this paper is organized as follows: In Section~\ref{sec:Problem_Statement}, we state the problem and our method proposed for the forecasting of the number of active cases. In Section~\ref{sec:Features}, we present the feature selection methods and the parameter optimization for each method. In Section~\ref{sec:Forecasting}, we present the implementation of the considered forecasting methods. In Section~\ref{sec:Results}, we present our results on the forecasting of the number of active cases in COVID-19 pandemic. In Section~\ref{sec:Conclusions}, we present our conclusions. 

\section{Statement of the Problem}
\label{sec:Problem_Statement}

In this section, we describe the forecasting problem for the active cases in COVID-19 outbreak. We aim to examine the generalization ability of the machine learning based forecasters for identifying their predictive powers on the COVID-19 data. To this end, we first analyze the effects of the different features on the forecasting of active cases and then select the important features that increase the forecasting accuracy. Second, we design forecasting models that perform prediction of the number of active cases. Furthermore, we analyze the performance of the forecasters for different forecasting horizons in an increasing order, and we provide the most reliable forecasting horizon for this problem by means of an empirical analysis.

\subsection{System Design}
\label{sec:SystemDesign}

\begin{figure}[h] 
	\centering
	\includegraphics[scale=0.5]{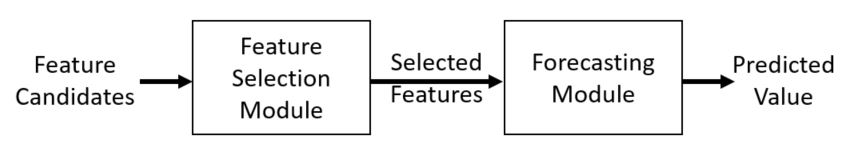}
	\caption{System Design of the Integrated Feature Selection - Forecasting for the Number of Active Cases in the COVID-19 Pandemic}
	\label{fig:SystemDesign}
\end{figure}

For the forecasting of the number of active cases, we design a system shown in Fig.~\ref{fig:SystemDesign} that consists of the Feature Selection module and the Forecasting module. The output of the system is the predicted value of each of the number of active cases for $1$- to $K$-day ahead forecasting. Furthermore, the detailed explanation of the methods that are used in the Feature Selection module and the Forecasting module are given in Section~\ref{sec:Features} and Section~\ref{sec:Forecasting}, respectively.

\subsection{Selection of the Important Features}
Since we know that there are many different features that may affect the spread of the COVID-19 outbreak, we analyze the features that we are able to access and select the feature subset. Each feature in this subset has important effects on the number of active cases. In order to improve the performance of the overall system, we perform the feature analysis combined with the forecasting module. That is, by using the feature selection methods in Section~\ref{sec:Features}, we select feature subset that achieves the best forecasting performance under the considered forecasting scheme. 

\subsection{Forecasting of the Active Cases}
In the forecasting problem, we aim to compute the future value of the active cases. To this end, we use machine learning models with supervised learning whose output is future value of the active cases at $K$th day. According to the best of the authors’ knowledge, there is no study that examines the maximum length of forecasting horizon that provides the forecasting within a reasonable accuracy for COVID-19 pandemic. In order to determine this horizon (which is the value of $K$), we forecast the total number of active cases for the increasing forecasting horizon length from $1$-day to $30$-days. We give the details of the machine learning methods that are used as the forecasting model and their input-output structures in Section~\ref{sec:Forecasting}

\section{Analysis and Selection of the Features}
\label{sec:Features}

In this section, we describe the methods for the selection of the relevant subset of features. For each country, first, we take past values of each of the following daily time series datasets: The number of total cases, the total number of deaths, and the total number of recovered patients. In order to convert the number of total cases to the number of active cases, which is actually the important value that will affect the control of the hospital facilities during the pandemic, we subtract the total sum of the deaths and recovered from the total cases. Then, in order to convert each of the total deaths and the total recovered into the per day basis, we take the difference between the samples. That is, the resulting three time series are the number of active cases, the number of deaths per day, and the number of recovered patients per day. 

Second, we have selected additional 36 different features that might affect the spread of COVID-19 and which are online available for all countries \cite{COVID19_merged, countryinfo}. Note that none of these features is not a time series data. The details of these features are given in Section~\ref{sec:Dataset}

In order to select the subset of these feature candidates, we apply three different feature selection methods: Iterative feature selection based on the pairwise correlation (PCorr) of each feature candidate pair (in short, correlation matrix), Recursive Feature Selection (RFS), and feature selection by using the Lasso regression (Lasso, in short). For each of the forecasting models given in Section~\ref{sec:Forecasting} for each value of $K$, we choose one of these feature selection methods by calculating the overall performance based on the cross-validation. 

\subsection{Feature Selection based on the Pairwise Correlation (PCorr)} 

In this method, first, for each feature, we calculate the correlation of this feature with each of the other features. After we complete the calculation of the pairwise correlation values for all of the feature candidate pairs, we compute the indices of the feature candidate pairs each of which whose correlation value is greater than the threshold value $p$ or less than $-p$. Third, from each pair of feature candidates, we eliminated a feature whose average correlation with other features is the greatest. Note that, we select the optimal value of $p$ by using an exhaustive search in the range of $[0.5, 0.99]$ with the increment of $0.01$.

In order to calculate the correlation matrix, we used the Pearson product-moment correlation, which is defined with the formula in Equation~\ref{eq:Pearson}. Each of the coefficients calculated by using this formula measures the strength and the direction of the linear relationship between two feature candidates.

\begin{equation}\label{eq:Pearson}
	\rho_{ij} =\frac{\frac{1}{L-1}\sum_{s=1}^{L}(x_i^s - \bar{x}_i)(x_j^s - \bar{x}_j)}{\sigma_{i} \sigma_{j}}, \qquad \forall i, j
\end{equation}

In Equation~\ref{eq:Pearson}, $\rho_{(i, j)}$ denotes the correlation value between the $i$th and the $j$th features. $x_i^s$ denotes the value of the $i$th feature candidate at sample $s$, $\bar{x}_i$ denotes the mean, and $\sigma_{i}$ denotes the standard deviation of the $i$th feature candidate. $L$ denotes the total number of samples. The numerator of this formula is the covariance between $i$th and $j$th features. 

\subsection{Recursive Feature Selection (RFS)}

The aim of the RFS is to shrink the set of the selected features in a recursive way. The RFS algorithm takes the set of feature candidates as the input. First, it trains an LR model with all of the feature candidates and keeps the coefficients of this LR. Note that we selected LR as the coefficient determination model in order to increase the generalization. The loop of the RFS searches for the optimal value of the desired number of features $N_\text{RFS}$ between $1$ and the total number $N$ of feature candidates. For each value of $N_\text{RFS}$, RFS works as follows: 1) It updates the input set by pruning $N - N_\text{RFS}$ features for which the LR coefficients are smaller than others. 2) The forecasting model is trained and tested with the resulting feature set by using cross-validation. 3) The mean of the test scores over the folds of the cross-validation is appended to the score list. When the end of the loop for $N_\text{RFS}$ is reached, the optimal value of $N_\text{RFS}^*$ is computed empirically. Finally, the resulting feature set is selected as to contain the first $N_\text{RFS}^*$ features that have the LR coefficients with the highest-values. 

\subsection{Feature Selection by using the Lasso Regression (Lasso)}

In order to select the subset of the feature candidates, we use the classical Lasso Regression with 5-fold cross-validation over the input sample set. For each fold of the cross-validation, we split data into the training and validation sets, then train the Lasso model on training set, test it on the test set and finally calculate the test score. As the best Lasso Regression model, we select the model that achieves the highest test performance over all of the models, each of which is trained in a cross-validation fold. Finally, we select the features each of whose Lasso coefficient is not equal to zero. 
\section{Forecasting of the Number of Active Cases}
\label{sec:Forecasting}

In this section, we describe how we forecast the future values of the number of active cases, and the detailed design of the forecasting module in Fig.~\ref{fig:SystemDesign}. 

In the forecasting module shown in Fig.~\ref{fig:SystemDesign}, we use $K$ different forecasters for $1$ to $K$ step ahead forecasting. Each of these forecasters is defined by its inputs, its output, and its internal model parameters. For each forecasting step $k$, we set the input of each forecasting model to the features that are selected as explained in Section~\ref{sec:Features}. We set the output of this forecaster to the value of the number of active cases at the $k$th step in the future, denoted by $\hat{x}_1[t+k]$. 

In order to forecast the future values of the number of active cases, we perform a comparative study with LR, MLP, and LSTM. We now describe the design and implementation of each of these models. 

\subsection{Linear Regression}

We have selected the well-known linear regression model as a benchmark forecaster. In the implementation of this model, we use the \textit{Linear Regression} module from \textit{scikit-learn} library \cite{scikitlearn}. The module fits a linear model with the coefficients to minimize the residual sum of squares between the observed targets in the dataset, and the predicted values by the linear approximation.

\subsection{Multi-Layer Perceptron}

We design an MLP model, which consists of two hidden layers. We let $n_l$ denote the number of neurons at hidden layer $l$. In order to find the local optimal architecture of the MLP model, we search for the values of $n_l$ for $l \in \{1, 2\}$ within the range of $[4, 32]$ for each integral power of two. We present the resulting architecture of the MLP model and compare the performances of these models in Section~\ref{sec:Results_Forecast}. Furthermore, we set the activation function of each neuron to $tanh$. In the implementation of the MLP model, we use the \textit{Keras} library in Python \cite{chollet2015keras}. 

\subsection{Long-Short Term Memory}

Our implementation of the LSTM model, which is coded by using \textit{Keras} library, consists of one lstm layer, two fully connected layers, and an output layer. We let $h_{lstm}$ denote the number of lstm units at the lstm layer and $h_{e}$ denote the number of neurons at each fully connected layer $e \in \{1, 2\}$. We exhaustively search for the local optimal values of $h_{lstm}$ and $h_{e}$ within the range of $[4, 32]$ for each integral power of two.

\section{Results}
\label{sec:Results}

\subsection{Dataset} 
\label{sec:Dataset}

In this paper, we have considered two different data domains. The first data domain is the time series data which consists of the number of active cases, the number of deaths and the number of recoveries for 71 different countries from 22th of January 2020 to 20th of July 2020. This domain contains one dataset collected from \cite{novel-corona-virus-2019-dataset}. The second data domain consists of two different datasets each of which includes different features that regard to each country. The first dataset in the second domain consists of 63 different features for 173 countries and is taken from \cite{COVID19_merged}. The second dataset in this domain consists of 58 different features for 194 countries and is taken from \cite{countryinfo}. 

We first got the intersection of all of the datasets with respect to the countries. Then, in the resulting dataset, we eliminated the features that are not available for all of the countries. Furthermore, we chose the subset of the country specific features, and we got 36 features. These country specific features are as follows: (1) latitude, (2) longitude, (3) population, (4) the number of people per kilometer square (in short, \emph{Density}), (5) urban population (in short, \emph{Urban-Pop}), (6) fertility, (7) median of the age (in short, \emph{Median-Age}), (8) average of the temperature between January 2020 and March 2020 (in short, \emph{Avg-Temperature}), (9) average of the humidity between January 2020 and March 2020 (in short, \emph{Avg-Humidity}), (10) the number of male children born per female giving birth (in short, \emph{Male-Birth}), (11-16) the number of males per female in overall (in short, \emph{MF}) and in the age groups 0-14 (in short, \emph{MF-14}), 15-25 (in short, \emph{MF-25}), 26-54 (in short, \emph{MF-54}), 55-64 (in short, \emph{MF-64}), and 65+ (in short, \emph{MF-65+}), (17) the percentage of the smokers (in short, \emph{Smokers}), (18) the number of beds in hospitals (in short, \emph{Bed-Capacity}), (19-21) the percentage of each of the female (in short, \emph{\% Female-Lung}), male (in short, \emph{\% Male-Lung}) and both female and male (in short, \emph{\% Lung}) that have lung diseases, (22) the death rate per 100000 caused by flu pneumonia (in short, \emph{Pneumonia-Death-100K}), (23) the binary flag that is equal to one for a country if the number of H1N1 cases is underestimated in 2009 for this country (in short, \emph{H1N1-Underestimate}), (24) the total number of confirmed cases per country during the H1N1 pandemic in 2009 (in short, \emph{H1N1-Confirmed}), (25) the total number of confirmed deaths caused by H1N1 during the H1N1 pandemic in 2009 (in short, \emph{H1N1-Deaths}), (26) the annual precipitation (in short, \emph(Annual-Precipitation), (27) the ratio of median property prices to the median familial disposable income (in short, \emph{Property-Affordability}), (28) the estimation \cite{HealthCareIndex} of the overall quality of the health care system (in short, \emph{Health-Care}), health care professionals, equipment, staff, doctors and cost, (29) the gross domestic product in 2019 (in short, \emph{GDP-2019}), (30) the health expenses in USD (in short, \emph{Health-Expenses}), (31) the health expenses per one million individuals (in short, \emph{Health-Expenses-1M}), (32) the limit of the number of person for gathering (in short, \emph{Gathering-Limit}), and (33-36) the number of days past between the date of the first confirmed case and each of the closing date of the non-essential public places (in short, \emph{Nonessential-Close-Days}), the starting date of the public gathering limitations (in short, \emph{Gathering-Limit-Days}), the closing date of the schools (in short, \emph{School-Close-Days}), and the closing date of the public places (in short, \emph{PublicPlace-Close-Days}).  

As a result, our dataset consists of $78$ features, where $42$ of them are the time series features during the COVID-19 pandemic, and $36$ of them are the general country related features. 

\begin{figure*} 
	\centering
	\includegraphics[scale=0.2]{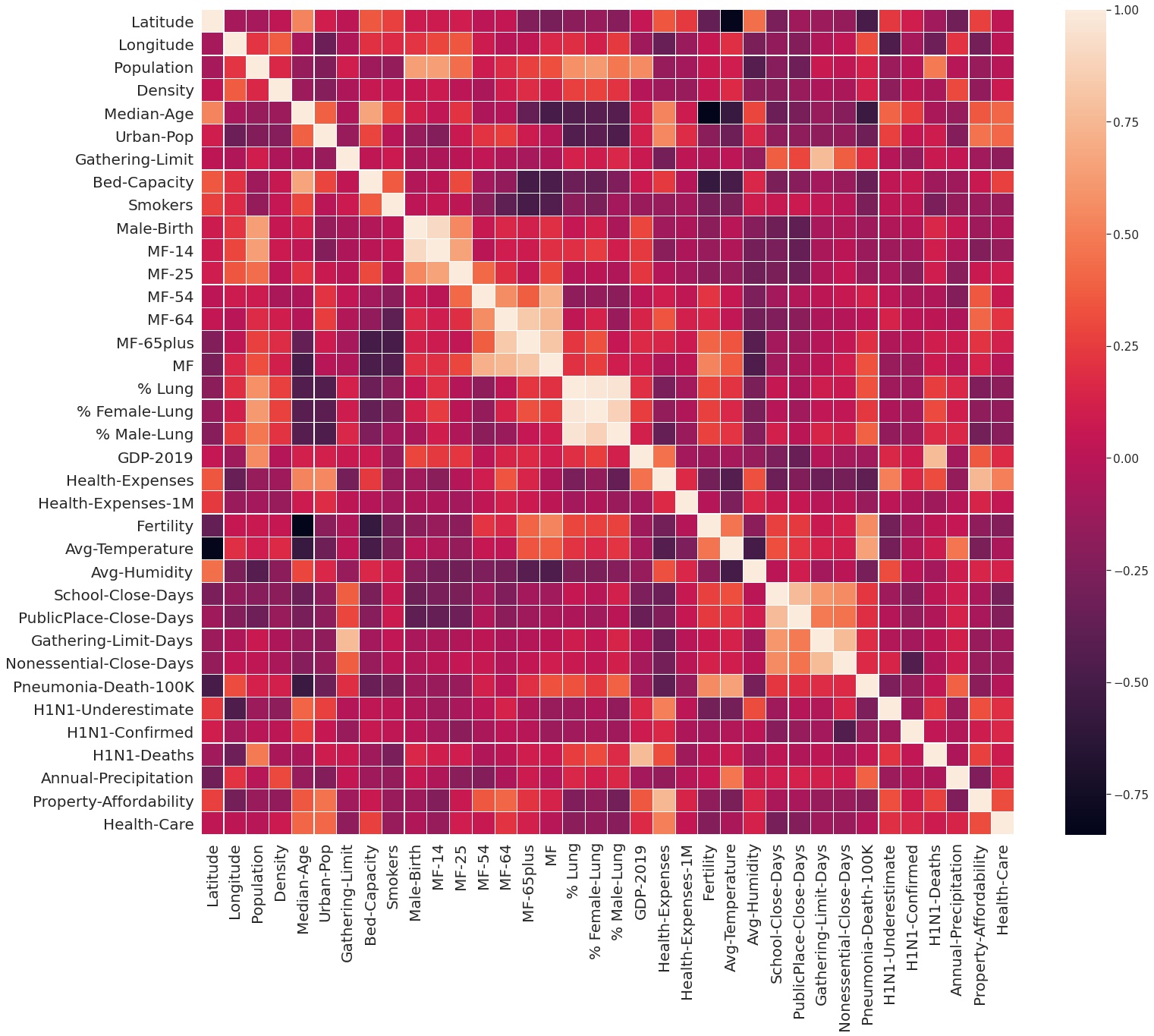}
	\caption{Heatmap of the pairwise pearson correlation for the feature candidates that are not time-series}
	\label{fig:heatmap}
\end{figure*}

%	https://www.numbeo.com/health-care/rankings_by_country.jsp

\subsection{Performance Evaluation by using 10-Fold Cross-Validation}

For each of the LR, MLP, and LSTM models, in order to measure the generalization ability of the model, we perform 10-fold Cross-Validation (CV). 

In each fold of the CV, we first split the dataset into the training set and test set randomly with ratios of $0.8$ and $0.2$, respectively. Second, we train the model on the training set and test it on the test set for the current fold. Third, we measure both of the training and test performance of the model by using the r$^2$ metric \cite{glantzslinker}.

In the training of the MLP and LSTM models, we use the ADAM algorithm as optimizer with the loss selected as the mean squared error (MSE). We set the parameters of the ADAM algorithm as follows: the initial learning rate to $0.001$, beta1 to $0.9$, beta2  to $0.999$. Furthermore, we set the batch size to $200$. During the training of MLP and LSTM models, we set the maximum number of epochs to $600$ for the early stopping that executes the training at the epoch where the training loss has not been decreasing for the last $30$ successive epochs. 

\subsection{Feature Selection Methods for the Forecasting of Active Cases}
\label{sec:Results_Feature}

In this subsection, for the forecasting problem of the active cases, we give the resulting parameters and the computation results for each of the PCorr, RFS, and Lasso feature selection methods. Furthermore, in Section~\ref{sec:Results_Forecast}, based on the r$^2$ score, we present the forecasting results for all of these feature selection methods for each forecasting model.

\subsubsection{Feature Selection based on the Pairwise Correlation (PCorr)} 

In Fig.~\ref{fig:heatmap}, we present the heatmap of the resulting pairwise pearson correlation for the feature candidates that are not time-series. In this figure, as the color of a pixel becomes whiter, the correlation value of the feature pair for the corresponding pixel increases. That is, the pixels each of whose color is very close to white and very close to black indicate the feature pairs that are highly correlated. For example, the Latitude of each country is inversely correlated with the Avg-temperature in that country, where the correlation value is $-0.82$. In addition, Median-Age and Fertility are inversely correlated with the correlation value $-0.84$. On the other hand, in Fig.~\ref{fig:heatmap}, we see that all of \% Lung, \% Male-Lung,\% Female-Lung features are highly correlated with correlation value above $0.87$.

%As a result of the exhaustive search, we found that the optimal values of the threshold $p$ is around $0.97$ for the forecasting steps $K = 1$ to $K = 15$, and it decreases gradually from $0.9$ to $0.83$ between $K=15$ and $K=30$. 
 
\subsubsection{Recursive Feature Selection (RFS)}
\begin{figure}[ht]
	\centering
	\includegraphics[scale=0.2]{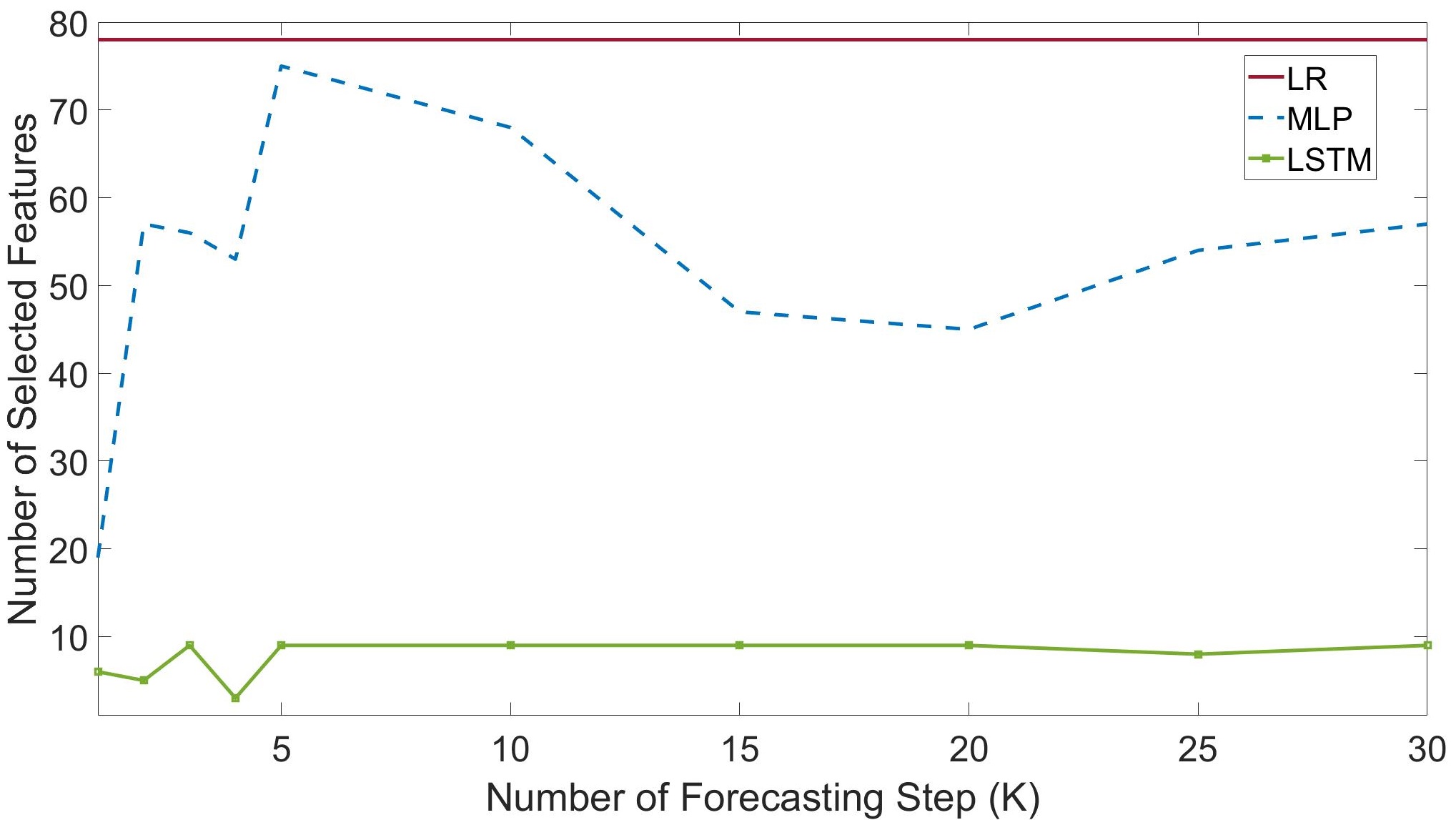}
	\caption{Comparison of the total number of selected features by RFS for each of LR, MLP, and LSTM forecasters}
	\label{fig:RFS}
\end{figure}

In Fig.~\ref{fig:RFS}, we present the optimal number of features $N_\text{RFS}$ that are selected by the RFS algorithm. In this figure, we see that the RFS selects all of the features for the LR forecaster. The reason is that the elimination of the features does not improve the LR performance any more. In addition, we see that the RFS selects only at most 9 features, which is the lowest number of features with respect the selected features for the other forecasters. The features that are selected for the LSTM model are the time series features with one exception, which is that the one of the selected features for the forecasting of day $30$ is the GDP-2019. Furthermore, we see that the numbers of selected features for LR and LSTM models do not change significantly with the increasing forecasting step; however, the number of selected features for the MLP forecaster varies with the increasing forecasting step.

%\subsubsection{Feature Transformation based on the Principal Component Analysis (PCA)}
%Principle Component Analysis used for Feature Map transformation and dimension reduction. In the implementation of this method, we used PCA library of \textit{scikit-learn}. As a tuning parameter, we feed the percentage of variance desired to conserve to the PCA algorithm. 

\subsubsection{Feature Selection by using the Lasso Regression (Lasso)}
\begin{figure}[ht]
	\centering
	\includegraphics[scale=0.2]{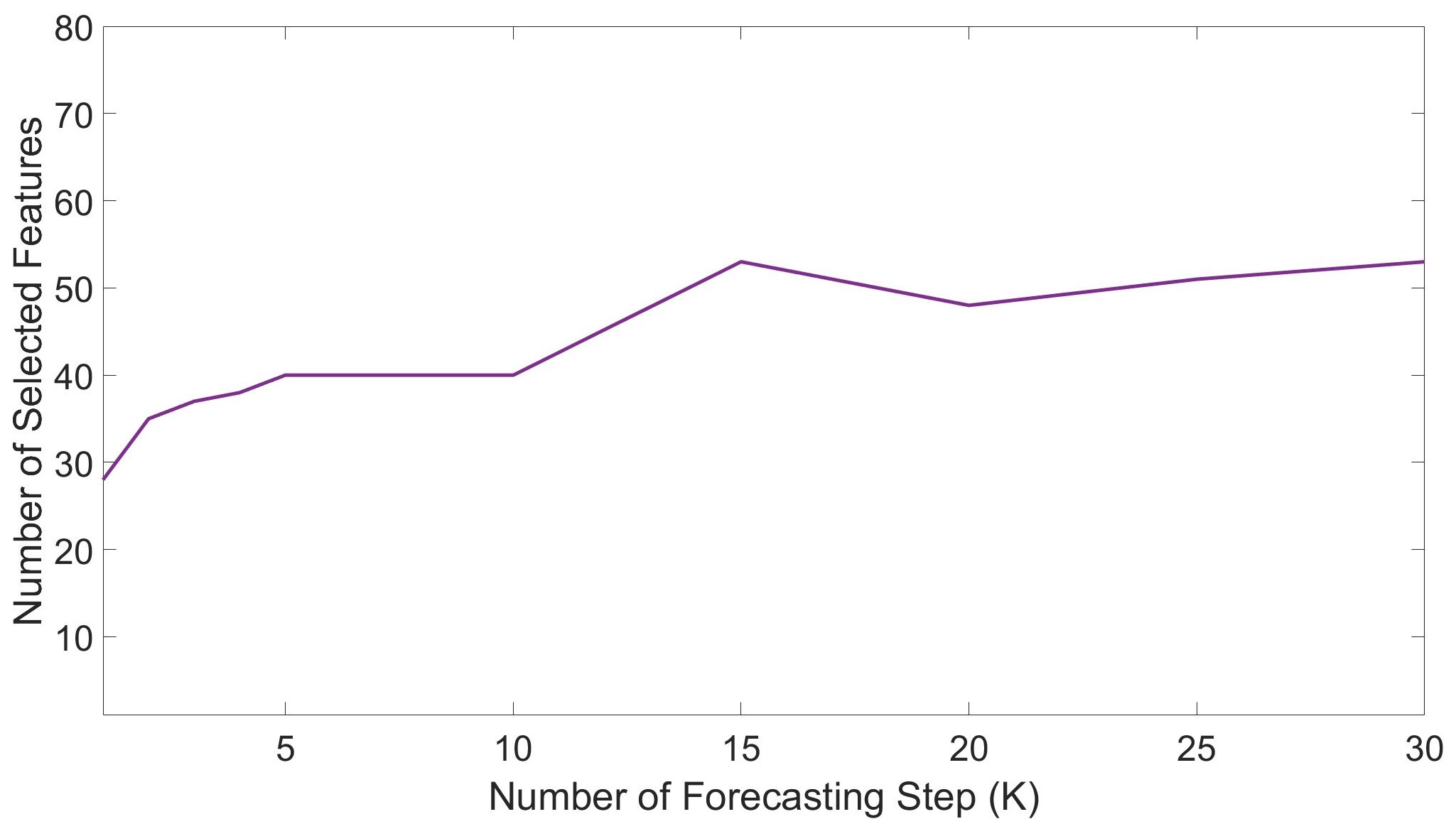}
	\caption{The total number of selected features by Lasso for the increasing number of forecasting step}
	\label{fig:Lasso}
\end{figure}

In Fig.~\ref{fig:Lasso}, we present the total number of selected features by Lasso for the increasing forecasting step. Note that the selected features by Lasso are the same for all forecasting models. We see that the Lasso tends to select more features as the number of forecasting step increases because the forecasting problem of determining the number of active cases becomes harder as the forecasting horizon widen. Furthermore, in the problem of forecasting of the near future, the number of active cases for the last day in the past has the best relationship with the desired output.  

\subsection{Forecasting Results for the Active Cases} 
\label{sec:Results_Forecast}

In this subsection, we discuss the predictability of the number of active cases for the COVID-19 pandemic. We also compare the forecasting performances of the LR, MLP and LSTM models.

\begin{figure}[h!]
	\centering
	\includegraphics[scale=0.22]{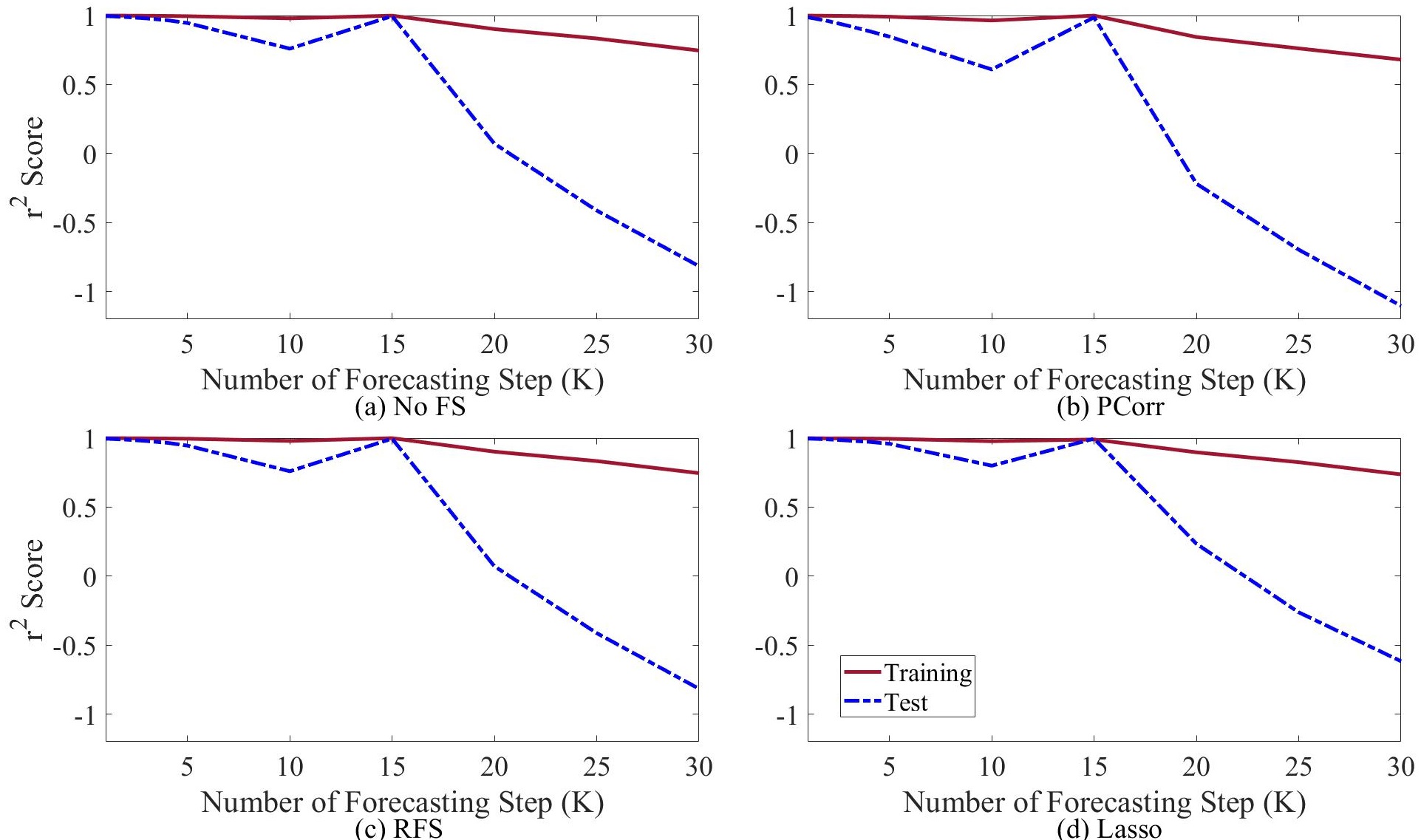}
	\caption{Forecasting performance of the Linear Regression for training and test sets under different feature selection methods.}
	\label{fig:LR}
\end{figure}

In Fig.~\ref{fig:LR}, we show the forecasting performance of the LR model under no feature selection (No FS), PCorr, RFS, and Lasso  with respect to r$^2$ metric. In this figure, for all of the feature selection methods, we see that the generalization gap between the training and the test performance of LR enlarges as the number of time step increases. In addition, the training performance (in r$^2$ metric) of the LR method is above $0.7$; however the test performance is under $0$ after $K = 20$. Even tough the LR is a linear model with high generalization ability, our results show that the LR model over fits the training data for the average of the CV folds. Furthermore, since the $99 \%$ of the COVID-19 patients develop symptoms and are hospitalized within $14$-days \cite{lauer2020incubation}, we see a significant performance improvement for $K = 15$ for all of the feature selection methods. That is, the LR model is able to capture the linear relationship between the past and the $15$-day ahead. Although the LR model cannot achieve a reasonable performance after $15$-days, it performs the best under Lasso.  

\begin{figure}[h!] 
	\centering
	\includegraphics[scale=0.22]{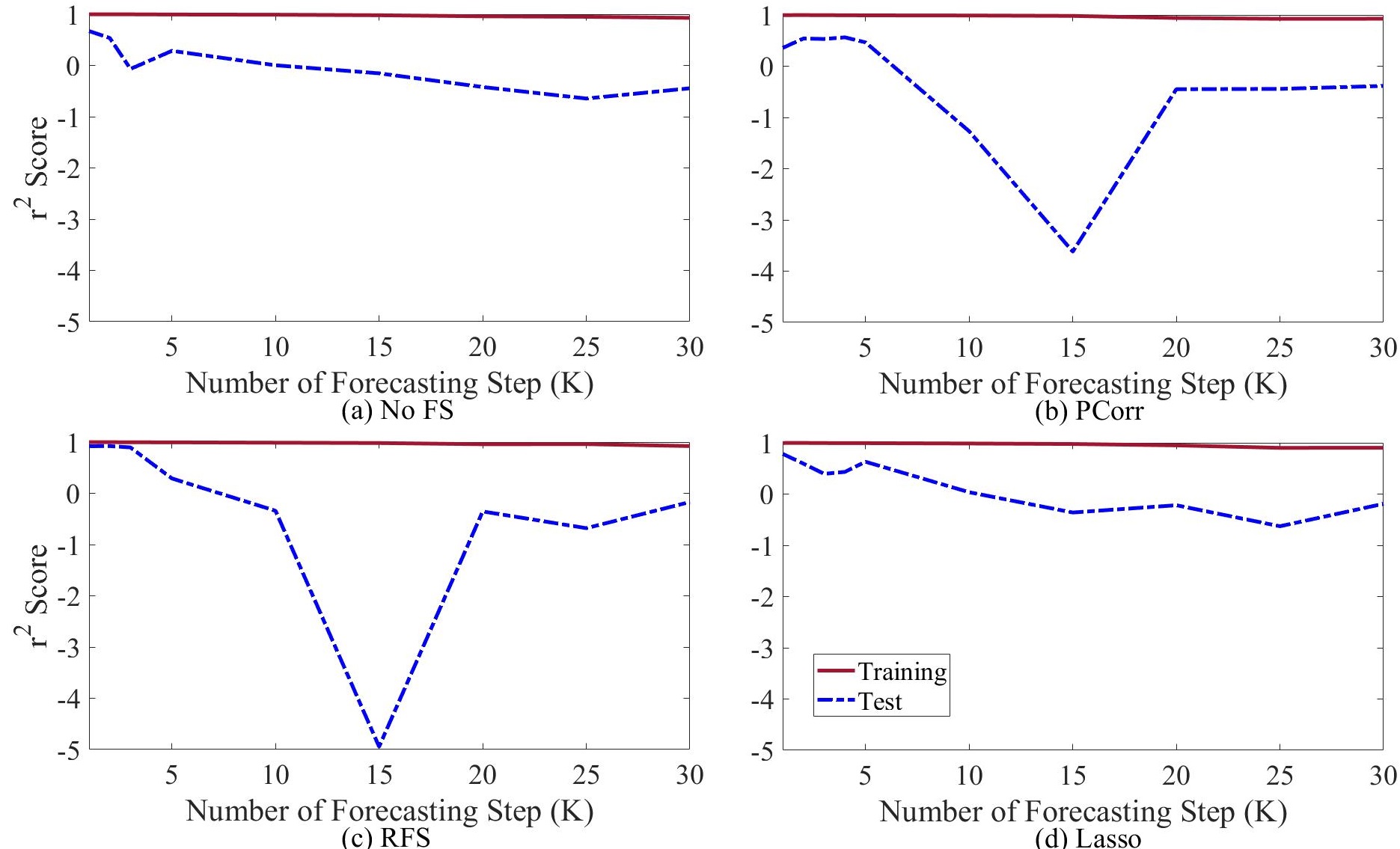}
	\caption{Forecasting performance of the MLP for training and test sets under different feature selection methods.}
	\label{fig:MLP}
\end{figure}

In Fig.~\ref{fig:MLP}, we present the CV results on both of the training and test sets for the MLP model under four different feature selection methods. First, we see that the mean of the training performance of MLP does not fall under $0.9$ for all of the feature selection methods. However, the test performance of the MLP model significantly decreases as the value of $K$ increases. That is, MLP, as a nonlinear model, highly over fits to the training data. Next, the MLP achieves its best r$^2$ performance under RFS up to $K = 5$, and its performances under No FS and under Lasso are comparable after $K = 5$.  

\begin{figure}[h!] 
	\centering
	\includegraphics[scale=0.22]{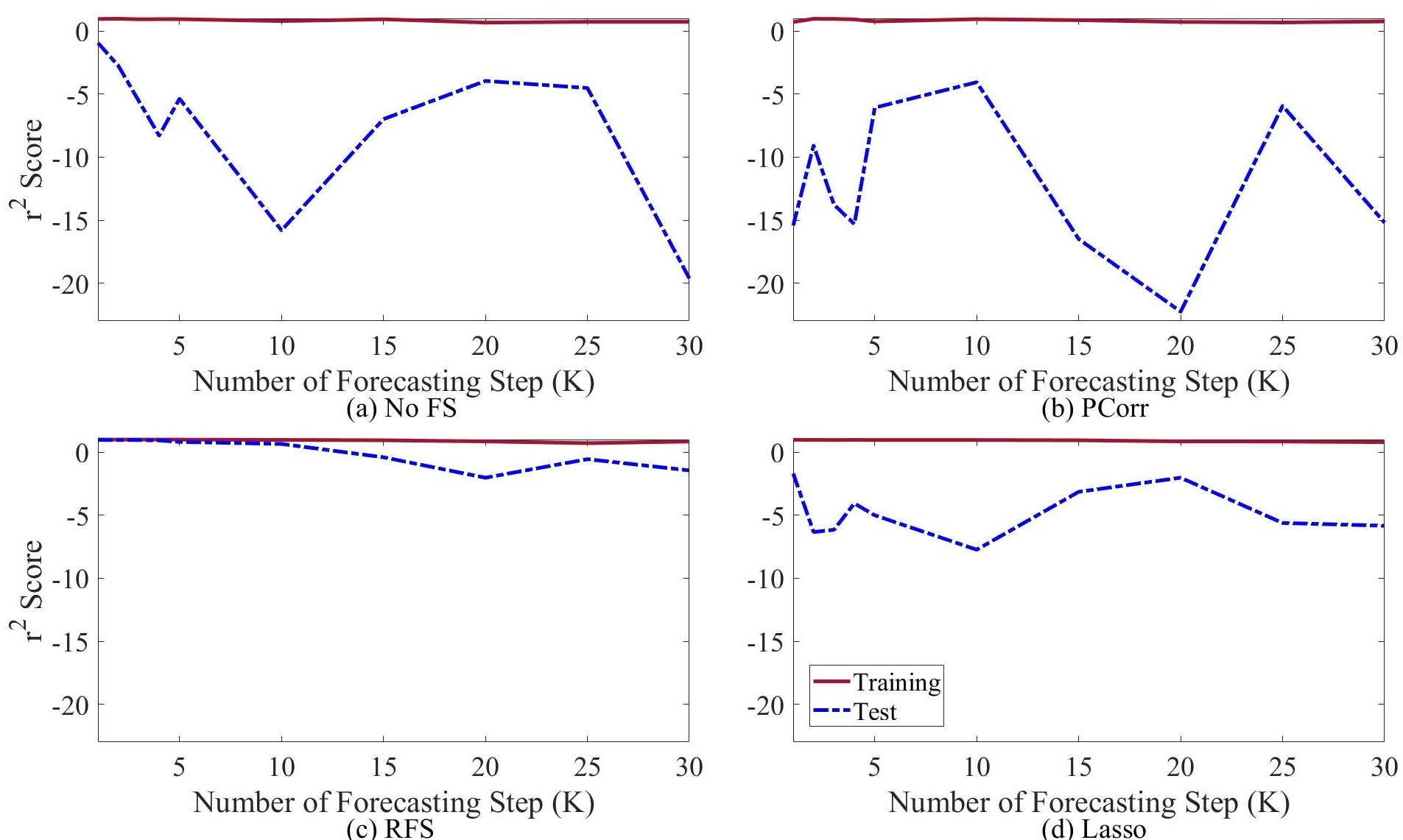}
	\caption{Forecasting performance of the LSTM for training and test sets under different feature selection methods.}
	\label{fig:LSTM}
\end{figure}

In Fig.~\ref{fig:LSTM}, we show both of the training and test r$^2$ performances of the LSTM model under four feature selection methods. For the LSTM model under each of the No FS, PCorr, RFS, and Lasso, since the training r$^2$ performance of the LSTM model does not fall under $0.73$, the test performance of that is under $0$ after $K = 15$. Furthermore, we see that RFS outperforms all of the other feature selection methods for the LSTM model. However, even the performance of the LSTM under RFS decreases to $-1.45$ r$^2$ value for $K=30$. 

\begin{figure}[h!] 
	\centering
	\includegraphics[scale=0.2]{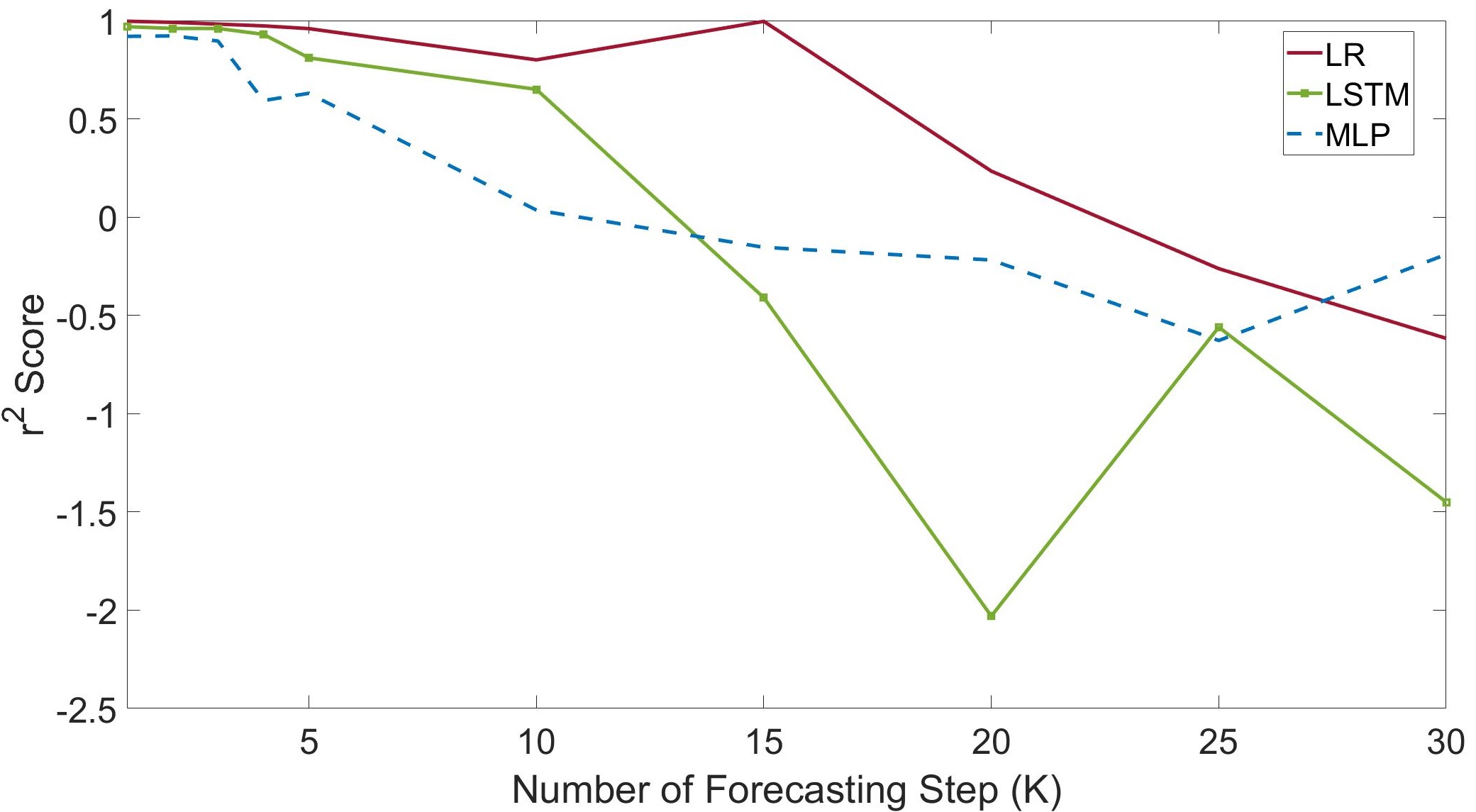}
	\caption{Comparison of the forecasting performance of LR, MLP, and LSTM under the best feature selection method for each value of $K$ with respect to the mean of the CV test scores.}
	\label{fig:All}
\end{figure}

In Fig.~\ref{fig:All}, we give the comparison of LR, MLP, and LSTM models each of which is applied together with the best performing feature selection method for each value of $K$. First, we see that only up to $K=3$, the r$^2$ performances of all models are higher than $0.9$. However, the MLP model significantly decreases at $K=3$, where this point is $K=4$ for the LSTM and $K=15$ for the LR. Second, we see that after $K=20$, there are no forecaster that achieves the r$^2$ score which is higher than $0$. It is concluded that since the number of samples for the forecasting problem of the number of active cases during COVID-19 pandemic is quite small to represent the feature space, we see that LR outperforms the other two models for all values of $K$, except $K=30$.

According to our results shown in Fig.~\ref{fig:LR}, Fig.~\ref{fig:MLP}, Fig.~\ref{fig:LSTM}, and Fig.~\ref{fig:All}, we see that due to the curse of small sample size, it is hard to forecast the number of active cases in COVID-19 outbreak with high generalization ability after $3$ days, except the $15$th day for which LR produces high prediction accuracy that might be due to the linear relation caused by the 14-day quarantine period applied to suspected persons.

\subsubsection{Forecasting of Active Cases on Extended Dataset}

In order to see the performance improvement with increasing sample size, we extended the dataset (which was collected from January 22 to July 20, 2020) with the number of active cases, the number of deaths and the number of recoveries for 71 different countries in COVID-19 pandemic until July 20, 2020. For this extended dataset, we repeat the methodology (in Section~\ref{sec:SystemDesign}) to generate the results in the rest of this section. 

\begin{figure}[h!] 
	\centering
	\includegraphics[scale=0.2]{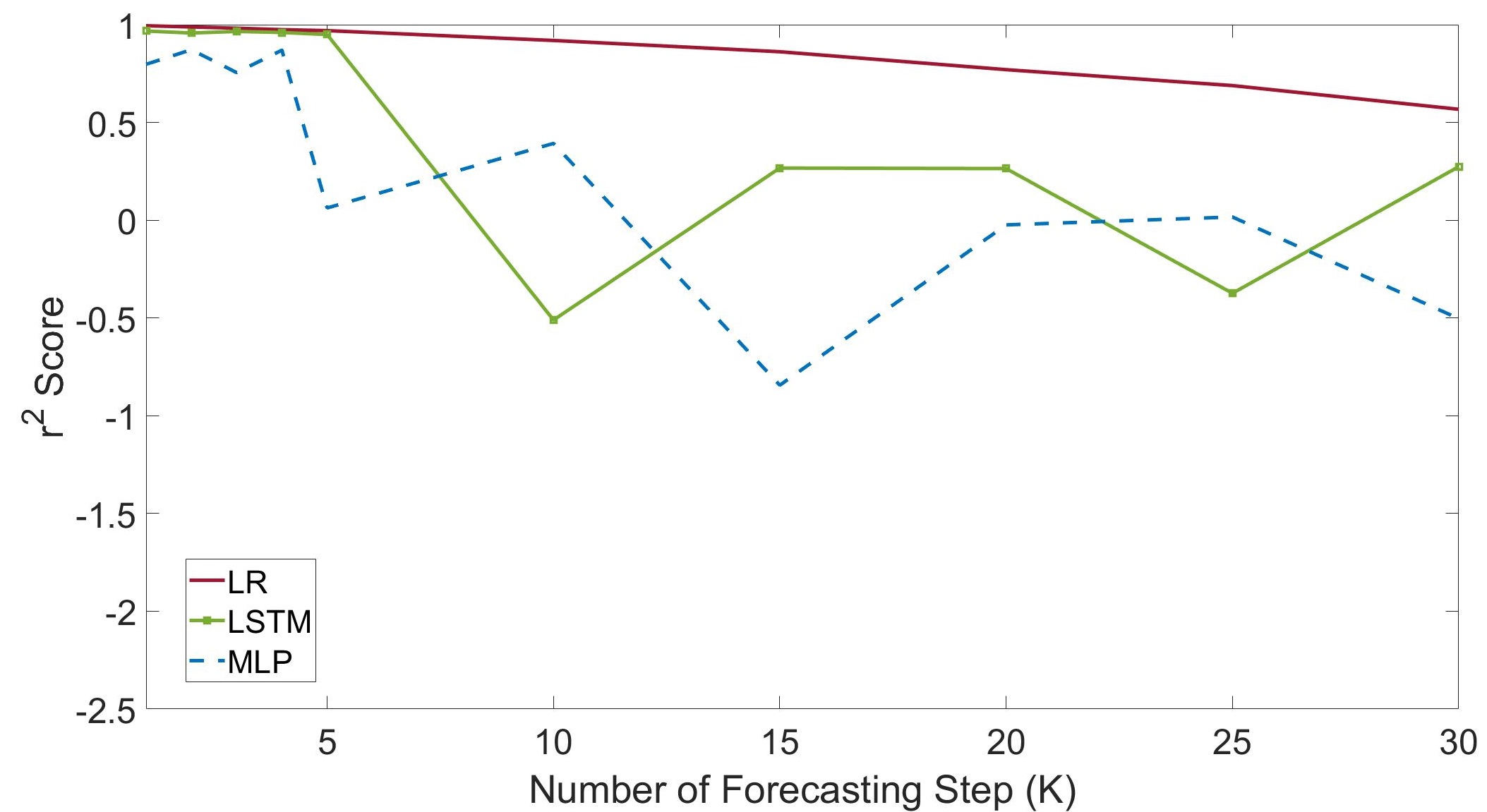}
	\caption{Comparison of the forecasting performance of LR, MLP, and LSTM under the best feature selection method for each value of $K$ with respect to the mean of the CV test scores on the extended dataset.}
	\label{fig:All_New}
\end{figure}

In Fig.~\ref{fig:All_New}, we display the r$^2$ performance of each forecasting scheme LR, LSTM, and MLP. We see that for each value of $K$, LR outperforms to both MLP and LSTM forecasters; however, even the performance of LR is around the $0.5$ for $K=15$. In addition, the r$^2$ performance of LSTM decreases after $K=5$, and that of MLP decreases significantly after $K=4$. Furthermore, due to the increased sample size of the dataset from Fig.~\ref{fig:All} to Fig.~\ref{fig:All_New}, we see that the performances of all forecasting schemes increase significantly for all value of $K$.

%Furthermore, for each time step ahead forecasting $K$, for the best architecture of MLP forecaster, the number of parameters is as follows: 2417, 1553, 1409, 1553, 1553, 3073, 3073, 3617, 5585, 3073. 

\subsubsection{Forecasting of Active Cases for Turkey}

\begin{figure*}[h!] 
	\centering
	\includegraphics[scale=0.37]{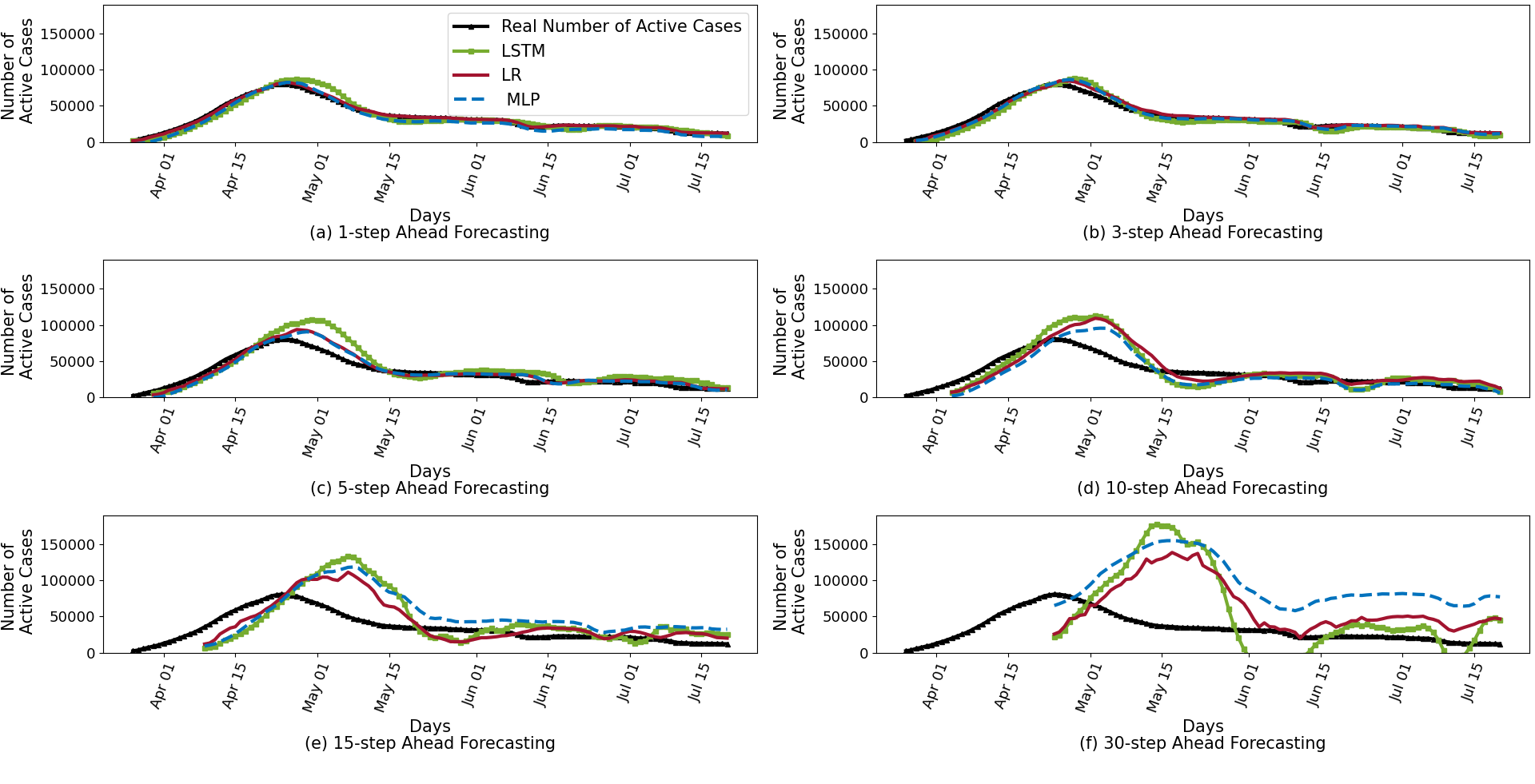}
	\caption{Comparison of the LR, MLP, and LSTM with respect to each of $1$, $3$, $5$, $10$, $15$ and $30$-step ahead forecasting of the number of active cases in Turkey between 26th of March 2020 and 20th of July 2020.}
	\label{fig:Forecasting_Turkey}
\end{figure*}

Now, in Fig.~\ref{fig:Forecasting_Turkey}, we present the forecasting results for the number of active cases for the increasing time step ahead forecasting in Turkey between 26th of March 2020 and 20th of July 2020. From Fig.~\ref{fig:Forecasting_Turkey}(a) to Fig.~\ref{fig:Forecasting_Turkey}(f), we respectively set the value of $K$ equal to $1$, $3$, $5$, $10$, $15$ and $30$. In Fig.~\ref{fig:Forecasting_Turkey}, for each value of $K$, we concatenate the $K$th-step ahead forecasting over the sliding windows with $1$-day sliding at each step.

In Fig.~\ref{fig:Forecasting_Turkey}(a), we see that the LR and MLP perform better than the LSTM forecaster, where LSTM is not able to forecast the number of active cases at around peak day. In this figure, except the days between 20th April and 10th May, all of the LR, MLP and LSTM models perform forecasting, which is close to real number of active cases. From Fig.~\ref{fig:Forecasting_Turkey}(a) to Fig.~\ref{fig:Forecasting_Turkey}(f), as the value of $K$ increases, we see that forecasting performances of all forecasting schemes decreases, and LR performs the closest forecasting to the real value of the number of active cases. Thus, in Fig.~\ref{fig:Forecasting_Turkey}(f), although the MLP forecasts close to real until 1st of May and LR forecasts close to real between 1st of June and 15th of June, we see that none of the forecasting models are able capture the general trend of the number of active cases and forecast the number of active cases for the peak day correctly for Turkey when $K=30$.  

\section{Conclusions}
\label{sec:Conclusions}

In this paper, we perform a study to determine the accurate forecasting horizon for the number of active cases in COVID-19 pandemic. To this end, we compare the performance of the Linear Regression (LR), Multi-Layer Perceptron (MLP), and Long-Short Term Memory (LSTM) for a variety of forecasting horizon lengths. Herein, the linear static model LR is chosen for its potentially high generalization ability. The most widely used static nonlinear neural network model MLP is preferred due to its powerful approximation property. The recurrent neural network LSTM is taken as the third benchmark model since it is the state-of-the-art model that is highly successful in capturing temporal relations in time series data. Considering the existence of limited number of samples only for COVID-19 pandemic, in order to achieve acceptable generalization ability for each of the three forecasters, we perform a feature selection to the input of the forecaster for reducing the model complexities. The forecaster under no feature selection (No FS) are then compared to the forecasters with the feature selection based on the Pairwise Correlation (PCorr), Recursive Feature Selection (RFS), and feature selection by using Lasso regression (Lasso), respectively. 

Our main conclusion is that the long term forecasting (in other words, prediction) of the number of active cases in COVID-19 pandemic is not possible with high test accuracy at least for the considered three benchmark models as a consequence of their poor generalization abilities under the very limited number of samples available, up to now, for the COVID-19 pandemic. This study is not conclusive. The other machine learning models such as 1-dimensional or multi-dimensional Convolutional Neural Networks may be applied for forecasting COVID-19 features such as active cases. However, all of these forecasting models will suffer from the small sample size problem. 

The study presented in this paper shows that the forecasting problem of the active cases might be solved by achieving the high performance and generalization ability up to $3$-days ahead only. In addition, this statement is also valid for the $15$th day ahead but only by using a linear model. Furthermore, even the best performing model is not able to perform better than fitting the mean of the data (which corresponds r$^2$ value equals to $0$) after $20$-days. 

\ifCLASSOPTIONcaptionsoff
  \newpage
\fi

\bibliography{References-COVID}
\vspace{-0.5in}
\begin{IEEEbiography}[{\includegraphics[width=1in,height=1.25in,clip]{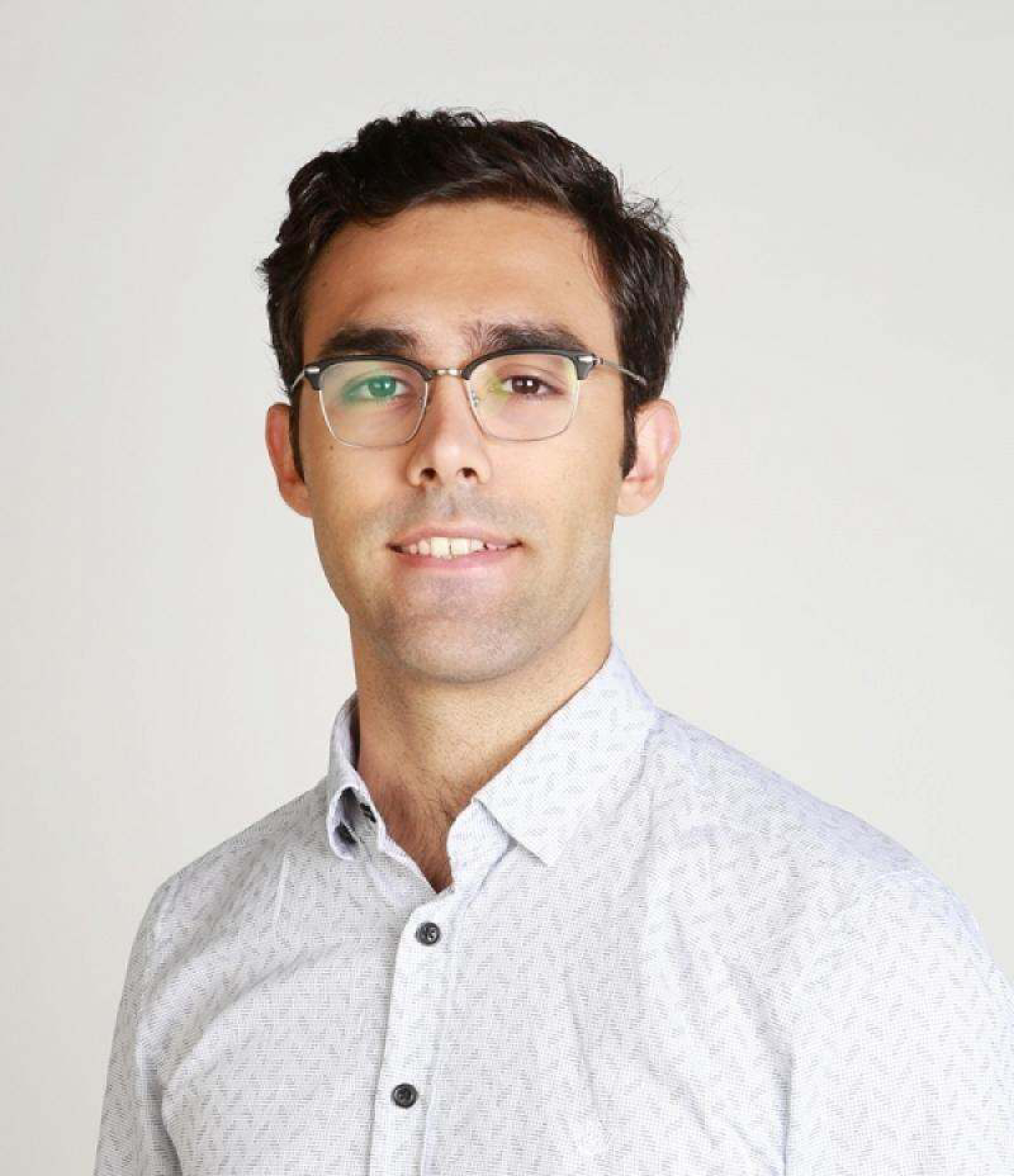}}]{Mert Nakıp} obtained his B.Sc. degree, with graduation rank \#1, from the Electrical-Electronics Engineering at Yaşar University (Izmir, Turkey) in 2018. His design of a multi-sensor fire detector via machine learning methods was ranked \#1 nationally at the Industry-Focused Undergraduate Graduation Projects Competition organized by TÜBİTAK (Turkish Scientific and Technological Research Council). He completed his M. Sc. thesis in Electrical-Electronics Engineering at Yaşar University (Izmir, Turkey) in 2020. His thesis focused on the application of machine learning methods to IoT and was supported by the National Graduate Scholarship Program of TÜBİTAK 2210C in High-Priority Technological Areas. He is currently a Research Assistant and a Ph.D. candidate at the Institute of Theoretical and Applied Informatics, Polish Academy of Sciences (Gliwice, Poland). 
\end{IEEEbiography}

\vspace{-0.5in}
\begin{IEEEbiography}[{\includegraphics[width=1in,height=1.25in,clip]{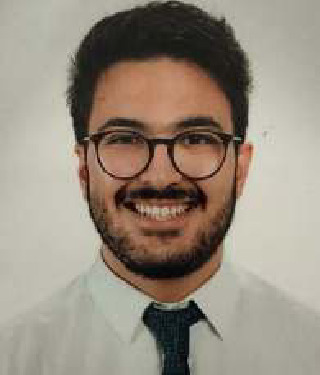}}]{Onur Çopur} received the B.Sc. degree (with Graduation Rank \#3) from the Department of Industrial Engineering, Yaşar University, İzmir, Turkey, in 2019. He is currently pursuing the M.Sc. degree in Data Science in Sapienza University of Rome. His research paper titled as "Periodic Route Optimization for FMCG Distributors", published as a chapter in the book called, "Proceedings of the International Symposium for Production Research 2019" by Springer and presented in the International Symposium for Production Research 2019 (ISPR2019) held at TU Wien. Vienna, Austria. 
\end{IEEEbiography}

\vspace{-0.5in}
\begin{IEEEbiography}[{\includegraphics[width=1in,height=1.25in,clip]{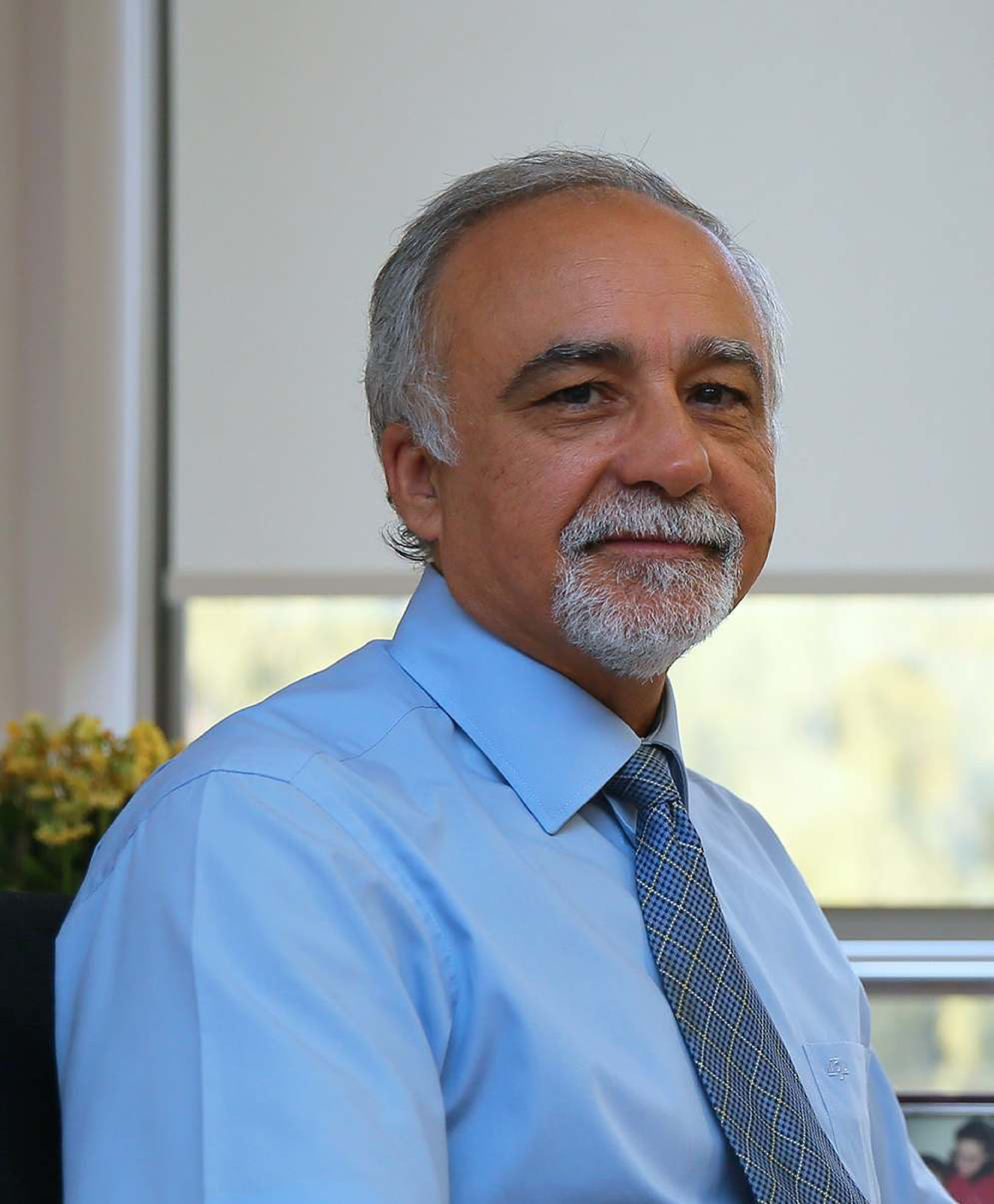}}]{Cüneyt Güzeliş} received the B.Sc., M.Sc., and Ph.D. degrees in electrical engineering from İstanbul Technical University, Istanbul, Turkey, in 1981, 1984, and 1988, respectively.
He is a Professor of electrical engineering and the Director of the Graduate School of Natural and Applied Sciences, Yaşar University, İzmir, Turkey. He was with İstanbul Technical University from 1982 to 2000, where he became a Full Professor. His expertise is in nonlinear circuits and systems, machine learning, and systems biology. He has published over 50 SCI-indexed journal papers with more than 900 SCI citations, six peer-reviewed book chapters, and more than 90 peer-reviewed conference papers. He has supervised 17 M.S. students and 14 Ph.D. students. He has served as the Dean of the Faculty of Engineering, Dokuz Eylül University, İzmir, and the Director of the Graduate School of Natural and Applied Sciences, İzmir University of Economics, İzmir. He has held visiting research and/or teaching positions with the University of California at Berkeley, Berkeley, CA, USA; Aachen University of Applied Sciences, Aachen, Germany; Paris-Est University, Champs-sur-Marne, France; and Paris-Nord University, Paris, France. He has participated in over 20 scientific research projects funded by the national and international institutions, such as the British Council and the French National Council for Scientific Research.
\end{IEEEbiography}

\end{document}